\documentclass[letterpaper]{article}
\pdfoutput=1
\usepackage{uai2018}
\usepackage[margin=1in]{geometry}

\usepackage{times}
\usepackage{natbib}

\usepackage{hyperref}
\usepackage{amssymb}
\usepackage{amsmath}
\usepackage{mathtools}
\usepackage{color}
\usepackage{graphicx}
\usepackage{dsfont}  
\usepackage{booktabs} 
\usepackage{multirow}

\usepackage{caption}  
\definecolor{atomictangerine}{rgb}{0.8, 0.2, 0.1}
\definecolor{turq}{rgb}{0.0, 0.5, 0.5}
\definecolor{darkturq}{rgb}{0.0, 0.4, 0.4}
\definecolor{bright}{rgb}{0.8, 0.1, 0}
\definecolor{darkgray}{gray}{0.3}
\definecolor{mahogany}{rgb}{0.6, 0.05, 0.05}
\definecolor{pink}{rgb}{1,0.05,0.6}
\definecolor{myblue}{rgb}{0.3,0.05,0.9}

\definecolor{olive}{rgb}{0.537, 0.627, 0.318}
\definecolor{green}{rgb}{0.22, 0.463, 0.114}
\definecolor{grey}{rgb}{0.4, 0.4, 0.4}
\definecolor{blue}{rgb}{0.435, 0.659, 0.863}
\definecolor{pink}{rgb}{0.761, 0.482, 0.627}

\newcommand{\tildeapprox}{{\raise.17ex\hbox{$\scriptstyle\sim$}}}

\renewcommand{\eqref}[1]{Eq.~(\ref{#1})}
\newcommand{\brackref}[1]{(\ref{#1})}

\newcommand{\lago}{LAGO}

\newcommand{\lagoK}{LAGO-K-Soft}
\newcommand{\Singletons}{LAGO-Singletons}
\newcommand{\Semantic}{LAGO-Semantic-Hard}
\newcommand{\SemanticSG}{LAGO-Semantic-Soft}

\newcommand{\reals}{\mathbb{R}}
\newcommand{\T}{\top}

\newcommand{\Z}{\mathcal{Z}}
\newcommand{\G}{\mathcal{G}}
\newcommand{\Gmk}{\Gamma_{m,k}}
\newcommand{\A}{\mathcal{A}}
\newcommand{\z}{z}
\newcommand{\X}{\mathcal{X}}
\newcommand{\Y}{\mathcal{Y}}
\newcommand{\K}{K}
\newcommand{\gkz}{g_{k,z}}
\newcommand{\gz}{\vec{g}_z}

\newcommand{\am}{a_{m}}
\newcommand{\ack}{\tilde{a_k}}

\newcommand{\Gk}{G_{k}}

\newcommand{\prob}{{p}}

\newcommand{\uuu}{{U}}
\newcommand{\vvv}{{V}}

\newcommand{\Smz}{\uuu_{m,z}}

\renewcommand\vec[1]{\mathbf{#1}}
\renewcommand{\a}{\vec{a}}
\newcommand{\x}{\vec{x}}

\newcommand\ignore[1]{}
\newcommand*{\medcup}{\mathop{\mathbin{\scalebox{1.5}{\ensuremath{\cup}}}}}%

\title{Probabilistic AND-OR Attribute Grouping for Zero-Shot Learning }

\author{ {\bf Yuval Atzmon$^*$ 
} \\
 $^*$Gonda Brain Research Center, \\
 Bar-Ilan University, 
 Israel  \\
 yuval.atzmon@biu.ac.il
 \And
{\bf Gal Chechik$^{*, **}$}  \\
$^{**}$Google AI, \\
 Mountain View, CA \\
 gal.chechik@biu.ac.il
}

\begin{document}

\maketitle

\begin{abstract}
In zero-shot learning (ZSL), a classifier is trained to recognize visual classes without any image samples. Instead, it is given semantic information about the class, like a textual description or a set of attributes. Learning from attributes could benefit from explicitly modeling structure of the attribute space. Unfortunately, learning of general structure from empirical samples is hard with typical dataset sizes.
    
Here we describe \lago{}\footnote{
A video of the highlights, and code is available at: \url{http://chechiklab.biu.ac.il/~yuvval/LAGO/} \\}, a probabilistic model designed to capture natural soft {\em and-or} relations across \textit{groups of attributes}. We show how this model can be learned  end-to-end  with a deep attribute-detection model. The soft group structure can be learned from data jointly as part of the model, and can also  readily incorporate prior knowledge about groups if available. The soft and-or structure succeeds to capture meaningful and predictive structures, improving the accuracy of zero-shot learning on two of three benchmarks.

Finally, \lago{} reveals a unified formulation over two ZSL approaches: DAP \citep{DAP} and ESZSL \citep{ESZSL}. Interestingly, taking only one singleton group for each attribute, introduces a \textit{new} soft-relaxation of DAP, that outperforms DAP by $\tildeapprox \text{40\%}$.


\end{abstract}


\begin{figure}[t]
    \centering
    \includegraphics[height=4.8cm, trim={3.4cm 1cm 2cm 2cm},clip]{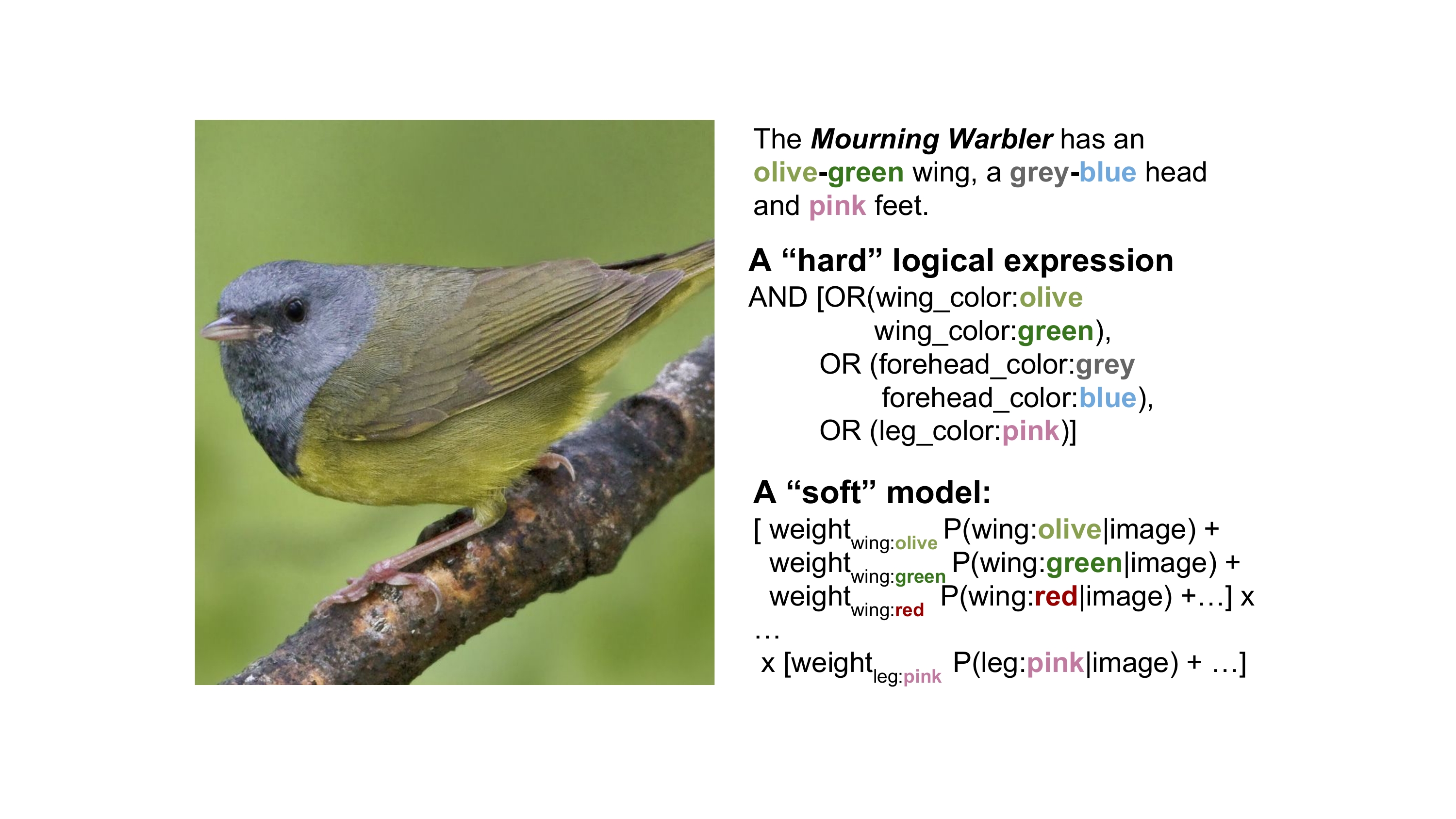} 
    \vspace{-10pt}
    \caption{Classifying a bird species based on attributes from \citep{CUB}.  
    The {\em Mourning Warbler} can be distinguished from other species by a combination of a grey head and olive-green underparts. Both human raters and machine learning  models may confuse  semantically-similar attributes like olive or green wings. These attribute naturally cluster into "OR" groups, where we aim to recognize this species if the wing is labeled as either green or olive. The \lago{} model (Eq. \ref{eq_prodsum}) weighs attributes detection, inferring classes based on within-group soft-OR and across-groups soft-AND. 
    In general, OR-groups include alternative choices of a property (\textit{wing\_color:\{\textcolor{mahogany}{red}, \textcolor{olive}{olive}, \textcolor{green}{green}\}}) and soft-OR allows to weigh down class-irrelevant attributes (here, \textit{wing:\textcolor{mahogany}{red}}).}
    \label{fig1}
\vspace{-10pt}
\end{figure}

\section{INTRODUCTION}
\label{sec_intro}
People can easily learn to recognize visual entities based on a handful of semantic attributes. For example, we can recognize a bird based by its visual features (long beak, red crown), or find a location based on a language description (a 2-stories brick town house). Unfortunately, when training models that use such semantic features, it is typically very hard to leverage semantic information effectively. With semantic features, the input space has rich and complex structure, due to nontrivial interactions and logical relations among attributes. For example, the color of petals may be red or blue but rarely both, while the size of a bird is often not indicative of its color.

Taking into account the semantics of features or attributes becomes crucial when no training samples are available. This learning setup, called zero-shot learning (ZSL) is the task of  learning to recognize objects from classes wihtout any image samples to train on. 
\citep{DAP,farhadi2009describing,palatucci2009zero,xianCVPR}. Instead, learning is based on semantic knowledge about the classes \citep{socher2013zero,elhoseiny2013write,berg2010automatic}, like in the case of attribute sharing \citep{DAP,lampert2014attribute}. Here, the training and test classes are accompanied by a set of predefined attributes, like "A Zebra is striped" or "A Hummingbird has a long bill", provided by human experts. Then, a classifier is trained to detect these attributes in images \citep{ferrari2008learning}, and test images are classified by detecting attributes and mapping to test classes based on the expert knowledge.

Broadly speaking, approaches to ZSL with attributes can be viewed as learning a compatibility function $f(Attr(image), Attr(class))$ between an attribute-based representation of an image and an attribute-based representation of classes \citep{ESZSL,ALE,DEVISE,SJE}. Here, the attributes of a class are provided by (possibly several) experts, the image attributes are automatically detected, and one aims to learn a scoring function that can find the class whose attributes are most compatible with an image.
Most ZSL approaches represent attributes as embedded in a ``flat'' space, Euclidean or Simplex, but flat embedding may miss important semantic structures. Other studies aimed to learn a structured scoring function, for example using a structured graphical model over the attributes \citep{WangBN}. Unfortunately, learning complex structures of probabilistic models from data requires large datasets, which are rarely available.

\begin{figure}[t]
    \centering
  \includegraphics[height=4.4cm, trim={7.4cm 3cm 3.5cm 3.5cm},clip]{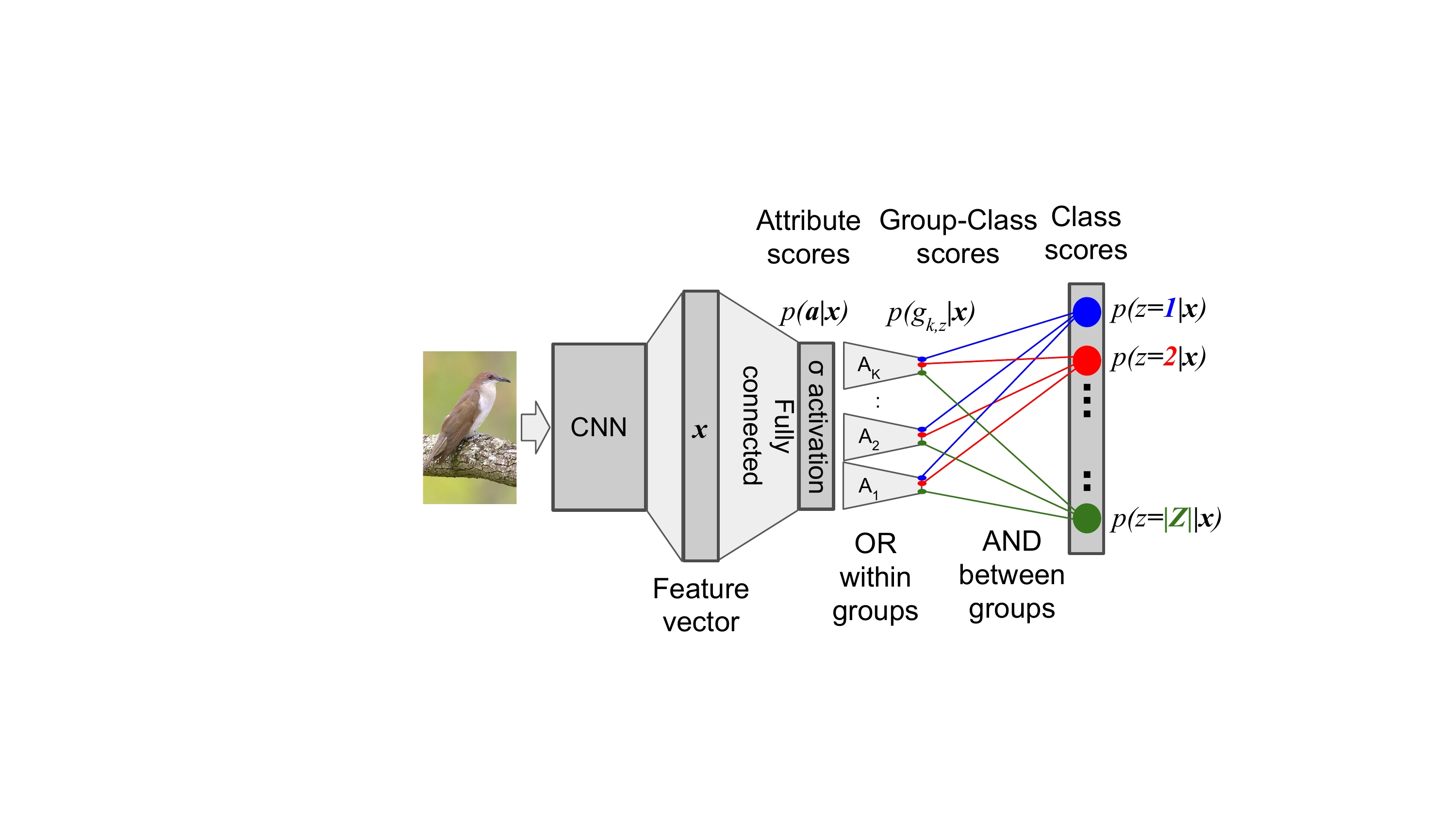} 
  \caption{\lago{} network architecture (Sec \ref{sec_zsl}). Image features are extracted by a deep ConvNet and fed to a FC sigmoid layer ($\sigma$) that outputs a prediction score $p(a_m|x)$ for each binary attribute $a_m$. Attribute scores are grouped into $K$ soft groups, and mapped to a set $\{\gkz\}$ of $K \times |Z|$ binary classifiers, according to a soft-OR \eqref{eq_gkz_cond_x}. Finally, a class is computed by the soft product of all group-scores for that class, approximating a conjunction (AND). 
    In the diagram, each colored circle represents a classifier score for a separate class. }
    \label{fig_illustration}
    \vspace{-10pt}
\end{figure}

Here we put forward an intermediate approach: We use training classes to learn a simple structure that can capture simple (soft) and-or logical relations among attributes. More concretely, after mapping an image to attributes, we aggregate attributes into groups using a soft OR (weighted-sum), and then score a class by taking a soft AND (product of probabilities) over group activations (Figure \ref{fig_illustration}). 
While the attributes are predefined and provided by experts, the soft groups are learned from the training data. 

The motivation for learning the and-or structure becomes clear when observing how attributes tend to cluster naturally into semantically-related groups. For example, descriptions of bird species in the CUB dataset include attributes like \{{\em wing-color:green}, {\em wing-color:olive}, {\em wing-color:red}\} \citep{CUB}. As another example, animal attributes in \citep{DAP} include \{{\em texture:hairless}, {\em texture:tough-skin}\}. In these two examples, the attributes are semantically related, and raters (or a classifier) may mistakenly interchange them, as evident by how Wikipedia describes the Mourning Warbler (Figure \ref{fig1}) as having ``olive-green underparts''. In such cases, it is natural to model attribute structure as a soft OR relation over attributes (``olive'' or ``green'') in a group (``underparts''). It is also natural to apply a soft AND relation across groups, since a class is often recognized by a set of necessary properties.

We describe \lago{}, "\textit{Learning Attribute Grouping for 0-shot learning}", a new zero-shot probabilistic model that leverages and-or semantic structure in attribute space. \lago{} achieves new state-of-the-art result on CUB and AWA2\citep{DAP}, and competitive performance on SUN \citep{SUN}. 
Interestingly, when considering two extremes of attribute grouping, \lago{} becomes closely related to two important ZSL approaches. First, in the case of a single group (all OR), \lago{} is closely related to ESZSL \citep{ESZSL}. At the opposite extreme where each attribute forms a singleton group, (all AND), \lago{} is closely related to DAP \citep{DAP}. \lago{} therefore reveals an interesting unified formulation over seemingly unrelated ZSL approaches.

Our paper makes the following novel contributions. We develop a new probabilistic model that captures soft logical relations over semantic attributes, and can be trained end-to-end jointly with deep attribute detectors. The model learns attribute grouping from data, and can effectively use domain knowledge about semantic grouping of attributes. We further show that it outperforms competing methods on two ZSL benchmarks, CUB and AWA2, and obtain comparable performance on another benchmark (SUN). Finally, LAGO provides a unified probabilistic framework, where two previous important ZSL methods approximate extreme cases of LAGO. 

\vspace{-5pt}
\section{RELATED WORK}
\label{sec_related}
\vspace{-5pt}
Zero-shot-learning with attributes attracted significant interest recently \citep{xianCVPR,recentAdvances}. One influential early works is {\em Direct Attribute Prediction} (DAP), which takes a Bayesian approach to predict unseen classes from binary attributes \citep{DAP}. In DAP, a class is predicted by the product of  attribute-classifier scores, using a hard-threshold over the semantic information of attribute-to-class mapping. DAP is related in an interesting way to \lago{}. We show below that DAP can be viewed as a hard-threshold special case of \lago{} where each group consists of a single attribute.

Going beyond a flat representation of attributes, several studies modeled structure among attributes. \citet{WangBN} learned a Bayesian network over attribute space that captures object-dependent and object-independent relationships.
\citet{Jiang2017} learned latent attributes that preserve semantics and also provide discriminative combinations of given semantic attributes.
Structure in attribute space was also used to improve attribute prediction: \citet{Jayaraman} leveraged side information about semantic relatedness of attributes in given groups and proposed a multi-task learning framework, where same-group attributes are encouraged to share low-level features.
In \cite{ParkANDOR, ParkANDOR2}, the authors propose an AND-OR grammar model \citep{ZhuANDORGrammer}, to jointly represent both the object parts and their semantic attributes within a unified compositional hierarchy. For that, they decompose an object to its constituent parts with a parse tree. In their model, the tree nodes (the parts) constitute an AND relation, and each OR-node points to alternative sub-configurations

Our approach resonates with \textit{Markov Logic Networks} (MLN), \citep{MLN2006} and {\em Probabilistic Soft Logic} (PSL), \citep{KimmigPSL, BachPSL}. It shares the idea of modeling domain knowledge using soft logical relations. Yet, LAGO formulation provides a complementary point of view:  (1) LAGO derivation reveals the probabilistic meaning of every soft weight of the logical relation \eqref{eq_gkz_cond_x}, and offers a principled way to set priors when deriving the soft logical expression. 
(2) The step-by-step derivation reveals which approximations are taken when mapping features to classes with soft logical relations. (3) Logical relations are incorporated in LAGO as part of an end-to-end deep network. PSL and MLN use Markov random field with logical relations as constraints or potentials.

The study of ZSL goes beyond learning with attributes \citep{SYNC, Tsai17Robust, morgadoCVPR17, Rohrbach11, AlHalah15, Zhang17, DSRL, REVISE, xu2017matrix, Li2018, zhang_kernel}. Recently, \citet{zhang_kernel} described a kernel alignment approach, mapping images to attribute space such that projected samples match the distribution of attributes in terms of a nonlinear kernel.
Another popular approach to ZSL learns a bi-linear compatibility function $F(\x,y; W) = \theta(\x)^{\T} W \phi(y)$ to match  visual information $\theta(\x)$ with semantic information $\phi(y)$ \citep{ESZSL,ALE,DEVISE,SJE}. In this context, most related to our work is ESZSL \citep{ESZSL}, which uses a one-hot encoding of class labels to define a mean-squared-error loss function. This allows ESZSL to have a closed-form solution where reaching the optimum is guaranteed. We show below that ESZSL is closely related to a special case of \lago{} where all attributes are assigned to a single group. 

The current work focuses on a new architecture for ZSL with attributes. Other aspects of ZSL,  including feature selection \citep{guo_attr_selection} and data augmentation \citep{CVAE1,CVAE2,xian_2018}, can improve accuracy significantly, but are orthogonal to the current work.

\vspace{-10pt}
\section{A PROBABILISTIC AND-OR MODEL}
\label{sec_zsl}

\paragraph{The Problem Setup:} Following the notations of \citep{DAP}, we are given a set of labeled training images $(\x_i \in \X, \z_i \in \Z)$ drawn from a distribution $D$.
Each image $\x$ is accompanied by a vector of binary attributes $\a\in\{True, False\}^{|A|}$, $\a=(a_1, \dots , \am, \dots, a_{|A|})$, where $\am = True$ if the image has attribute $\am$. We are also given a set of class "descriptions" in the form class-conditioned attribute distribution $p(\a|z)$. In practice, the descriptions are often collected separately per attribute ($m$), and only the marginals $p(a_m|z)$, $\forall m$ are available.

At training time, we aim to learn a classifier that predicts a class $z$ of an image $\x$ by first learning to predict the attributes $\a$, and then use $p(a_m|z)$, $\forall m$ to predict a class $z$ based on the attributes.

At inference time (the zero-shot phase), we are given images $\x$ from new unseen classes with labels $y \in \Y$, and together with their class descriptions $\prob(\am|y)$, $\forall m$. We similarly predict the class $y$ by first predicting attributes $\a$ and then use $p(a_m|z)$, $\forall m$ to predict a class $y$ based on the attributes.

\paragraph{Model Overview:}
\label{sec_overview}
The LAGO model (Figures \ref{fig1}, \ref{fig_illustration}) learns a soft logical AND-OR structure over semantic attributes. It can be viewed as a concatenation of three mapping steps $\X \rightarrow \A \rightarrow \G \rightarrow \Z$. First, \textbf{attribute predictions:} an image $\x$ is mapped to a vector of attribute detection probabilities $f^1_W: \X \rightarrow \A$, $\A = [0,1]^{|A|}$. The mapping parameters $W$ determine the weights of the attribute detectors and are learned from labeled training data. Second, \textbf{weighted-OR group scores:} Attribute probabilities are mapped to $K$ groups. Each group calculates a class-dependent weighted-OR over the $|Z|$ classes $f^2_{\uuu,\vvv}:\A \rightarrow \G$, $\G =[0,1]^{K\times|Z|}$. The mapping parameters $\uuu$ are the distributions $p(\a|z)$ provided with each class;  The mapping parameters $\vvv$ determine how attributes are grouped and are learned from data.
Last, \textbf{soft-AND group conjunction}: Per-group scores are mapped to class detection probabilities by a soft-AND, approximating group conjunction. 
$f^3:\G \rightarrow \Z$, $\Z = [0,1]^{|Z|}$.
The parameters $W,\vvv$ are learned jointly to minimize a regularized loss with a regularizer $R$:
\begin{equation}
    \min_{W,\vvv} \quad loss(f^3(f^2_{\uuu,\vvv}(f^1_W(\x_i))),  z_i) + R(W, \vvv)\quad.
\vspace{-5pt}
\end{equation}

The key idea in the proposed approach is to define a layer of binary classifiers $\gkz$, each evaluating a class $z$ based only on a subset $\Gk\subset\A$ of attributes. For example, for bird-species recognition, one classifier may detect a {\em Mourning Warbler} based on wing colors and another based on  bill shapes. In this example, each of the classifiers output the probability $p(\gkz=True|\a)$ that the image has a {\em Mourning Warbler}, but based on different subsets of attributes. The partition of attributes to subsets is shared across classes, hence with $K$ subsets we have $K \times |Z|$ binary classifiers. We also define $\a_k$ to be the vector of attributes detections for $\Gk$, $\a_k \in \{T,F\}^{|\Gk|}$. For clarity, we first derive the algorithm for groups that are fixed (not learned) and hard (non-overlapping). Section  \ref{sec_soft} then generalizes the derivation to soft learned groups.

Consider now how we compute $p(\gkz\!=\!T|\x)$. According to the Markov property, it equals $\sum_{\a_k}p(\gkz|\a_k)p(\a_k|\x)$, but computing this sum raises several challenges. First, since the number of possible patterns in a group grows exponentially with its size, the summation becomes prohibitively large when attribute groups are large. Second, estimating $p(\gkz|\a_k)$ from data may also be hard because the number of samples is often limited. Finally, description information is often available for the marginals only, $(z,a_m)$, rather than the full distribution $(z, \a_k)$. We now discuss a model that addresses these constraints.

\paragraph{The Within-Group Model $\A \rightarrow \G$:}
\label{sec_factor}
We now show how one can compute $p(\gkz|\x)$ efficiently by treating attributes within group as obeying a soft OR relation. As discussed above, OR relations are in good agreement with how real-world classes are described using hard semantic attributes, because a single property (a group like {\em beak-shape}) may be mapped to several semantically-similar attributes ({\em pointy}, {\em long}).

Formally, we first define a complementary attribute per group, $\ack = \left(\medcup_{m \in \Gk} a_m\right)^c$, handling the case where no attributes are detected or described, and accordingly define $\Gk' = \Gk \medcup \ack$. We then use the identity $\prob(A) = \prob(A,B) + \prob(A,B^c)$ to partition $\prob(\gkz\!=\!T|\x)$ to a union (OR) of its contributions from each of its attributes. Specifically, $\prob(\gkz\!=\!T|\x) = \prob(\gkz\!=\!T, \medcup_{m \in \Gk} a_m=T|\x) +  \prob(\gkz\!=\!T,\ack=T |\x)$ $=\prob(\medcup_{m \in \Gk'}(\gkz\!=\!T, a_m=T)|\x)$.
Using this equality and approximating
attributes within a group as being mutually exclusive, we have 
\begin{equation}
    \label{eq_disj_mux}
    \prob(\gkz\!=\!T|\x) \approx \sum_{m \in \Gk'} \prob(\gkz\!=\!T, a_m=T|\x) .
\end{equation}
To rewrite this expression in terms of class descriptions $p(a_m|z)$ we take the following steps. First, the Markov chain $\X \rightarrow \A \rightarrow \G$ gives  $\prob(\gkz\!=\!T,\am |\x)$ $=$ $\prob(\gkz\!=\!T |\am)\prob(\am|\x)$.
Second, we note that by the definition of $\gkz$, $p(\gkz|\a_k) = p(z|\a_k)$, because $\gkz$ is the classifier of $z$ based on $\a_k$. This yields $\prob(\gkz\!=\!T, \am) = \prob(z,\am)$ by marginalization. Applying Bayes to the last identity gives $\prob(\gkz\!=\!T|\am) = {\prob(\am|z)\prob(\gkz\!=\!T)}/{\prob(\am)}$ (more details in Supplementary Material \ref{sec_am_cond_gkz_supp}). 
Finally, combining it with the expression for $\prob(\gkz\!=\!T,\am |\x)$ and with \eqref{eq_disj_mux} we can express $\prob(\gkz\!=\!T|\x)$ as 

\vspace{-10pt}
\begin{multline}
    \label{eq_gkz_cond_x}
    \prob(\gkz\!=\!T|\x) {} \approx \\ \prob(\gkz\!=\!T)\sum\limits_{m \in \Gk'} \frac{\prob(\am=T|z) }{\prob(\am=T)} \prob(  \am=T | \x).
\end{multline}

\paragraph{Conjunction of Groups $\G\rightarrow\Z$: }
\label{sec_jointly}
Next, we derive an expression of the conditional probability of classes $\prob(z|\x)$ using soft-conjunction of group-class classifiers $\gkz$. Using the Markov property $\X \rightarrow \A \rightarrow \G \rightarrow \Z$, and denoting ${g_{1,z} \dots g_{\K,z}}$ by $\gz$, we can write \\
$\prob(z|\x)\! = \! \sum_{\gz} \prob(z|\gz)\prob(\gz|\x)$. We show on Supplementary \ref{sec_derivation_eq_conj_supp}, that 
making a similar approximation as in DAP \citep{DAP}, for groups instead of attributes, yields \eqref{eq_conj}: 
$\prob(z|\x) \approx \prob(z)\prod_{k=1}^K \frac{\prob(\gkz=T|\x)}{\prob(\gkz=T)}$.
Combining it with \eqref{eq_gkz_cond_x}, we conclude
\begin{equation}
    \label{eq_prodsum}
    \prob(z|\x) {} \approx \,\,\,\,\prob(z) \prod\limits_{k=1}^{\K}\Big[\!\!\sum_{m \in \Gk'} \!\!\frac{\prob(\am\!=\!T|z) }{\prob(\am\!=\!T)} \prob(\am\!\!=\!\!T|\x)  \Big] .
\end{equation}

\vspace{-10pt}
\subsection{SOFT GROUPS:}
\label{sec_soft}
The above derivation treated attribute groups as hard: deterministic and non-overlapping. We now discuss the more general case where attributes are {\em probabilistically} assigned to groups.

We introduce a soft group-membership variable $\Gmk = \prob({m \in \Gk'})$, yielding a soft version of \eqref{eq_gkz_cond_x}
\vspace{-5pt}
\begin{multline}
    \label{eq_soft_gkz_cond_x}
    \prob(\gkz\!=\!T|\x) {} \approx \\ \prob(\gkz\!=\!T)  \sum_{m = 1}^{|\A|} \Gmk\frac{\prob(\am=T|z) }{\prob(\am=T)} \prob(  \am=T | \x),
\end{multline}
\vspace{-15pt}

\noindent where each row of $\Gamma$ represents a distribution over $K$ groups per attribute in the simplex $\Delta^{K}$. Hard grouping is a special case of this model where all probability mass is assigned to a single group for each row of $\Gamma$. The full derivation is detailed in Supplementary Material (\ref{sec_derviation_soft_groups_supp}). 

\vspace{-10pt}
\subsection{LEARNING}
\vspace{-5pt}
\label{sec_learning}

\label{sec_learning_suppl}
\lago{} has three sets of parameters learned from data. 

First, a matrix $W$ parametrizes the mapping $f^1_W: \x \rightarrow [0,1]^{|A|}$ from image features to attribute detection probabilities $\prob(\am|\x)$. This mapping is implemented as a fully-connected layer with sigmoid activation over image features extracted from ResNet-101.

Second, a matrix $U$, where its entry $\Smz$ parametrizes the class-level description $\prob(\am|z)$. When attribute ratings are given per image, we estimate $\uuu$ using maximum likelihood from co-occurrence data over attributes and classes. 

Third, a matrix $\vvv_{|A| \times K}$ parametrizes the soft group assignments $\Gamma_{|A| \times K}$, such that each row $m$ maintains $\Gamma_{(m, :)} = \text{softmax}(\zeta \vvv_{(m, :)})$, where  $\zeta\in\reals^+$ is a smoothing coefficient. This parametrization allows taking arbitrary gradient steps over $\vvv$, while guaranteeing that each row of $\Gamma$ corresponds to a probability distribution in the simplex $\Delta^{K}$. 

Since $W$ and $\vvv$ are shared across all classes, they are learned over the training classes and transferred to the test classes at (zero-shot) inference time. They are learned end-to-end by applying cross-entropy loss over the outputs of \eqref{eq_prodsum} normalized by their sum across classes (forcing a unit sum of class predictions). As in \citep{ESZSL}, the objective includes two regularization terms over $W$: A standard $L_2$ regularizer $||W||^2_{Fro}$ and a term  $||W\uuu||^2_{Fro}$, which is equivalent for an ellipsoid Gaussian prior for $W$. For the "\textit{\SemanticSG{}}" learning-setup (Section \ref{sec_compared_methods}) we introduce an additional regularization term $||\Gamma_{(\vvv)} - \Gamma_{(\vvv_{SEM})}||^2_{Fro}$, pushing the solutions closer to known  semantic hard-grouping $\Gamma_{(\vvv_{SEM})}$. Finally, we optimize the loss:
\begin{multline}
    \label{eq_loss}
    L(W,\uuu,\vvv,Z,A,X) {} =  \text{CXE}_{\prob(z|\x; W,\uuu,\vvv)}(X, Z) + \\ \alpha \text{BXE}_{\prob(a|\x; W)}(X, A)+  \beta ||W||^2_{Fro} +  \lambda ||WS||^2_{Fro}  + \\ \psi ||\Gamma_{(\vvv)} - \Gamma_{(\vvv_{SEM})}||^2_{Fro},
\end{multline}
where CXE is the categorical cross-entropy loss for $\prob(z|\x)$, BXE  is the binary cross-entropy loss for $\prob(\a|\x)$, $X, Z$ and $A$ denote the training samples, labels and attribute-labels. Per-sample attribute labels are provided as their empirical mean per class.  In practice, we set $\alpha=0$ (See Section \ref{sec_experimental_setup_components}) and cross-validate to select the values of $\beta$, $\lambda$  and $\psi$ when relevant.

\ignore{ 
We optimize the following loss: (detailed in Supplementary \ref{sec_learning_suppl})
\begin{multline}
    \label{eq_loss}
    L(W,\uuu,\vvv,Z,A,X) {} =  \text{CXE}_{\prob(z|\x; W,\uuu,\vvv)}(X, Z) + \alpha \text{BXE}_{\prob(a|\x; W)}(X, A)+ \\ \beta ||W||^2_{Fro} +  \lambda ||WU||^2_{Fro}  + \psi ||\Gamma_{(\vvv)} - \Gamma_{(\vvv_{SEM})}||^2_{Fro}
\end{multline}
where $W$ parametrizes the mapping $\X \rightarrow \A$;  $\vvv_{|A| \times K}$ parametrizes the soft group assignments $\Gamma_{|A| \times K}$; 
$\Smz$ is the class-level description $\prob(\am|z)$, estimated with maximum likelihood from co-occurrence of attributes and classes;
CXE is categorical cross-entropy loss for $\prob(z|\x)$; BXE  is binary cross-entropy loss for $\prob(\a|\x)$;
$L_2$ regularizers $||W||^2_{Fro}$,   $||W\uuu||^2_{Fro}$ as in \citep{ESZSL}; $||\Gamma_{(\vvv)} - \Gamma_{(\vvv_{SEM})}||^2_{Fro}$ is a prior of semantic hard-grouping of $\Gamma$ for "\textit{\SemanticSG{}}" (Section \ref{sec_compared_methods});
$X, Z, A$ denote training samples, labels and attribute-labels;  $\alpha, \beta, \lambda, \psi$ are hyper-params. 
}

\vspace{-5pt}
\subsection{INFERENCE}
\vspace{-5pt}
\label{sec_inference}
At inference time, we are given images from new classes $y \in \Y$. As with the training data, we are given semantic information about the classes in the form of the distribution $p(\am|y)$. 
In practice, we are often not given that distribution directly, but instead estimate it using maximum likelihood from a set of labeled attribute vectors. 

To infer the class of a given test image $\x$, we plug $p(\am|y)$ estimates instead of $p(\am|z)$ in \eqref{eq_prodsum}, and select the class $y$ that maximizes \eqref{eq_prodsum}.

\vspace{-10pt}
\subsection{DAP, ESZSL AS SPECIAL CASES OF \lago{}}
\vspace{-5pt}
\lago{} encapsulates similar versions of two other zero-shot learning approaches as extreme cases: DAP \citep{DAP}, when having each attribute in its own singleton group ($K=|A|$), and  ESZSL \citep{ESZSL}, when having one big group over all attributes ($K=1$). 

Assigning each single attribute $\am$ to its own singleton group reduces \eqref{eq_prodsum} to \eqref{soft_dap} (details in Supplementary \ref{sec_dap_eszsl_supp}).
This formulation is closely related to DAP. When expert annotations $\prob(\am=T|z)$ are thresholded to $\{0,1\}$ and denoted by $a^z_m$, \eqref{soft_dap} become the DAP posterior \eqref{dap}. This makes the singletons variant a \textbf{new} soft relaxation of DAP. 

At the second extreme (details in Supplementary \ref{sec_dap_eszsl_supp}), all attributes are assigned to a single group, $K\!=\!1$. Taking a uniform prior for $\prob(z)$ and $\prob(\am)$, and replacing $\prob(\am\!\!=\!\!T|\x)$ with the network model $\sigma(\x^\T W)$, transforms \eqref{eq_prodsum} to
$\prob(z|\x) \!\!\propto \!\!\!\!  \sum_{m=1}^{|A|} \sigma(\x^\T W) \prob(\am\!\!=\!\!T|z)$.
Denoting $\Smz\!\!=\!\!\prob(\am\!\!=\!\!T|z)$, this formulation reveals that at the extreme case of $K\!=\!1$, \lago{} can be viewed as a non-linear variant  that is closely related to ESZSL: $Score(z|\x)\!\!=\!\!\x^\T W U$, with same entries $\Smz$. 

\section{EXPERIMENTS}
\vspace{-5pt}
\label{sec_compared_methods}
Fair comparisons across ZSL studies tends to be tricky, since not all papers use a unified evaluation protocol. To guarantee an "apple-to-apple" comparison, we follow the protocol of a recent meta-analysis by \citet{xianCVPR} and compare to the leading methods evaluated with that protocol: 
\textbf{DAP} \citep{DAP}, 
\textbf{ESZSL} \citep{ESZSL},
\textbf{ALE} \citep{ALE},
\textbf{SYNC} \citep{SYNC},
\textbf{SJE} \citep{SJE},
\textbf{DEVISE} \citep{DEVISE},
\textbf{Zhang2018} \citep{zhang_kernel}. Recent work showed that data augmentation and feature selection can be very useful for ZSL \citep{CVAE1, CVAE2, xian_2018, guo_attr_selection}. Since such augmentation are orthogonal to the modelling part, which is the focus of this paper, we do not use them here. 

\vspace{-5pt}
\subsection{DATASETS}
\label{sec_datasets}
\vspace{-5pt}
We tested \lago{} on three datasets: CUB, AWA2 and SUN. First, we tested \lago{} in a fine-grained classification task of bird-species recognition using CUB-2011 \citep{CUB}. CUB has 11,788 images of 200 bird species and a vocabulary of 312 binary attributes (wing-color:olive), derived from 28 attribute groups (wing-color). Each image is annotated with attributes generated by one rater. We used the class description $p(\am|z)$ provided in the data. The names of the CUB attributes provide a strong prior for grouping (wing-color:olive, wing-color:red, \dots $\rightarrow$ wing-color:\{olive, red, \dots\}). 


The second dataset, Animals with Attributes2 (AWA2),  \citep{xian_awa2} consists of 37,322 images of 50 animal classes with pre-extracted feature representations for each image. 
Classes and attributes are aligned with the class-attribute matrix of  \citep{Osherson91,Kemp2006}.
We use the class-attribute matrix as a proxy for the class description $p(\am|z)$, since human subjects in \citep{Osherson91} did not see any image samples during the data-collection process. As a prior over attribute groups, we used the 9 groups proposed by  \citep{AWAgroups, Jayaraman} for 82 of 85 attributes, like \textit{texture:\{furry, hairless, \dots \}} and \textit{shape:\{big, bulbus, \dots\}}.
We added two groups for remaining attributes: \textit{world:\{new-world, old-world\}}, \textit{smelly:\{smelly\}}. 

As the third dataset, we used SUN \citep{SUN}, a dataset of complex visual scenes, having 14,340 images from 717 scene types and 102 binary attributes from four groups. 


\vspace{-5pt}
\subsection{EXPERIMENTAL SETUP}
\vspace{-5pt}
\label{sec_experimental_setup_components}
We tested four variants of \lago{}:  \\
\textbf{(1) \Singletons{}:} The model of \eqref{eq_prodsum} for the extreme case using  $K=|A|$ groups, where each attribute forms its own hard group. \\
\textbf{(2) \Semantic{}:} The model of \eqref{eq_prodsum} with hard  groups determined by attribute names. As explained in Section \ref{sec_datasets} . \\
\textbf{(3) \lagoK{}:} The soft model of Eqs. \\ \ref{eq_prodsum}-\ref{eq_soft_gkz_cond_x}, learning $K$ soft group assignments with $\Gamma$ initialized uniformly up to a small random perturbation. $K$ is a hyper parameter with a value between $1$ and the number of attributes. It is chosen by cross-validation. \\
\textbf{(4) \SemanticSG{}:} The model as in \lagoK{}, but the soft groups $\Gamma$ are initialized using the dataset-specific semantic groups assignments. These are also used as the prior $\Gamma_{SEM}$ in \eqref{eq_loss}. 

Importantly, to avoid implicit overfitting to the test set, we used the validation set to select a single best variant, so we can report only a single prediction accuracy for the test set. For reference, we provide detailed test results of all variants in the Supplementary Material, Table \ref{table_results_all_variants} .  

\begin{figure}[t]
    \centering
    \includegraphics[width=7.5cm, trim={0cm 0cm 0cm 0cm},clip]{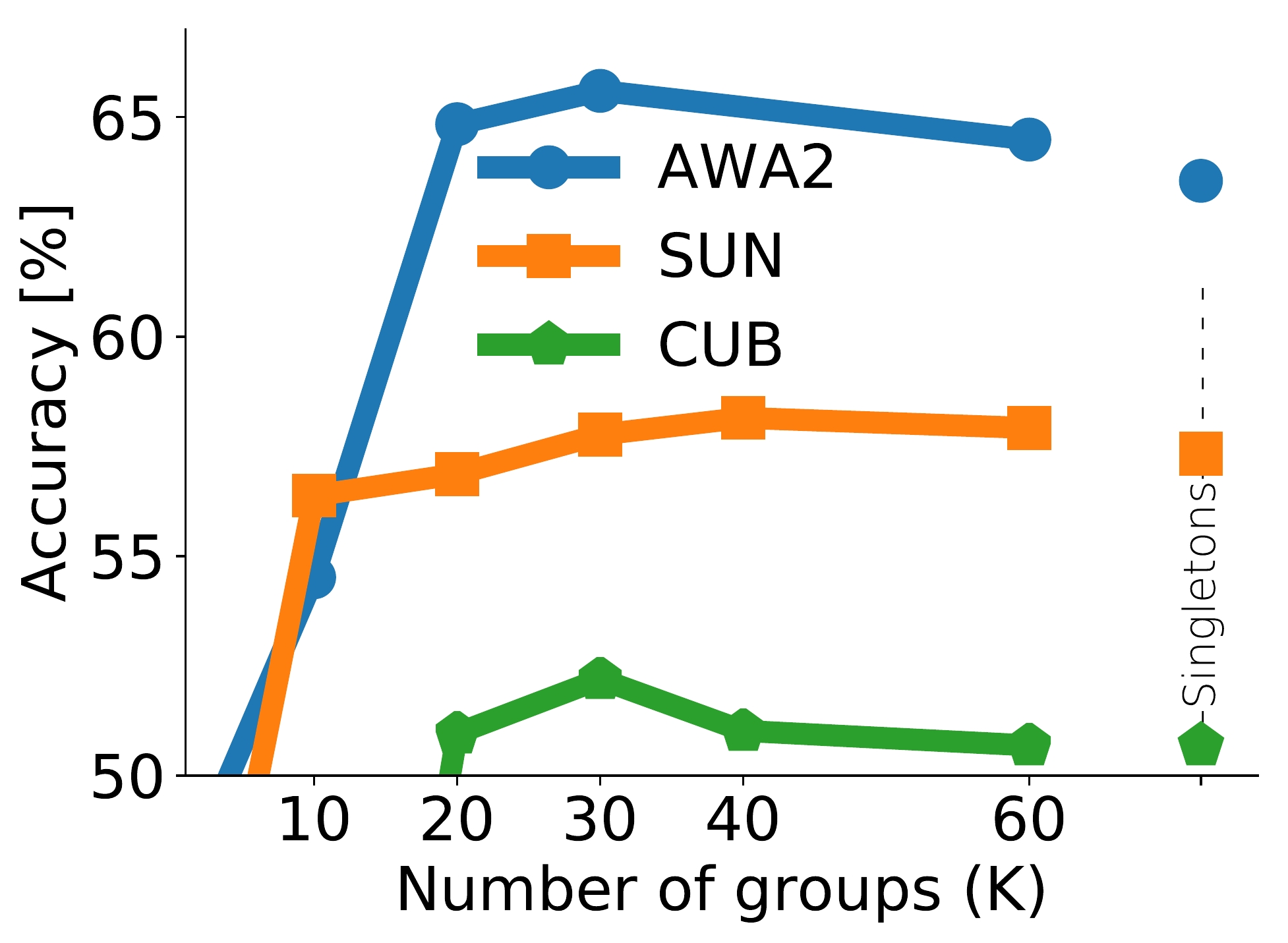} 
    \caption{Learning K soft-group assignments: Validation-set accuracy for different number of groups ($K$) for the \lagoK{} variant and of the \Singletons{} baseline. Here, no prior information is given about the groups, and the model successfully learns groups assignments from data, and better than the group-naive \Singletons{} baseline.}
    \label{fig_acc_vs_K_on_val}
\vspace{-5pt}
\end{figure}

To learn the parameters $W, \vvv$, we trained the weights with cross entropy-loss over  outputs (and regularization terms) described in section \ref{sec_learning}. \textbf{In the hard-group case,} we only train $W$, while keeping $\vvv$ fixed. We sparsely initialize $\vvv$ with ones on every intersection of an attribute and its hard-assigned group and  choose a high constant value for $\zeta$ ($\zeta=10$). Since the rows of $\Gamma$ correspond to attributes, it renders each row of $\Gamma$ as a unit mass probability on a certain group. \textbf{In the soft-group case,} we train $W, \vvv$ alternately per epoch, allowing us to choose different learning rate for $W$ and $\vvv$. For \lagoK{}, $\vvv$ was initialized with uniform random weights in [0, 1e-3], inducing a uniform distribution over $\Gamma$ up to a small random perturbation.\textbf{ For \SemanticSG{},} we initialized $\vvv$ as in the hard-group case, and we also used this initialization for the prior $\vvv_{SEM}$ \eqref{eq_loss}.

\begin{figure}[t]
    \centering
    \includegraphics[width=8.5cm,  trim={0cm 0.3cm 0cm 0cm},clip]{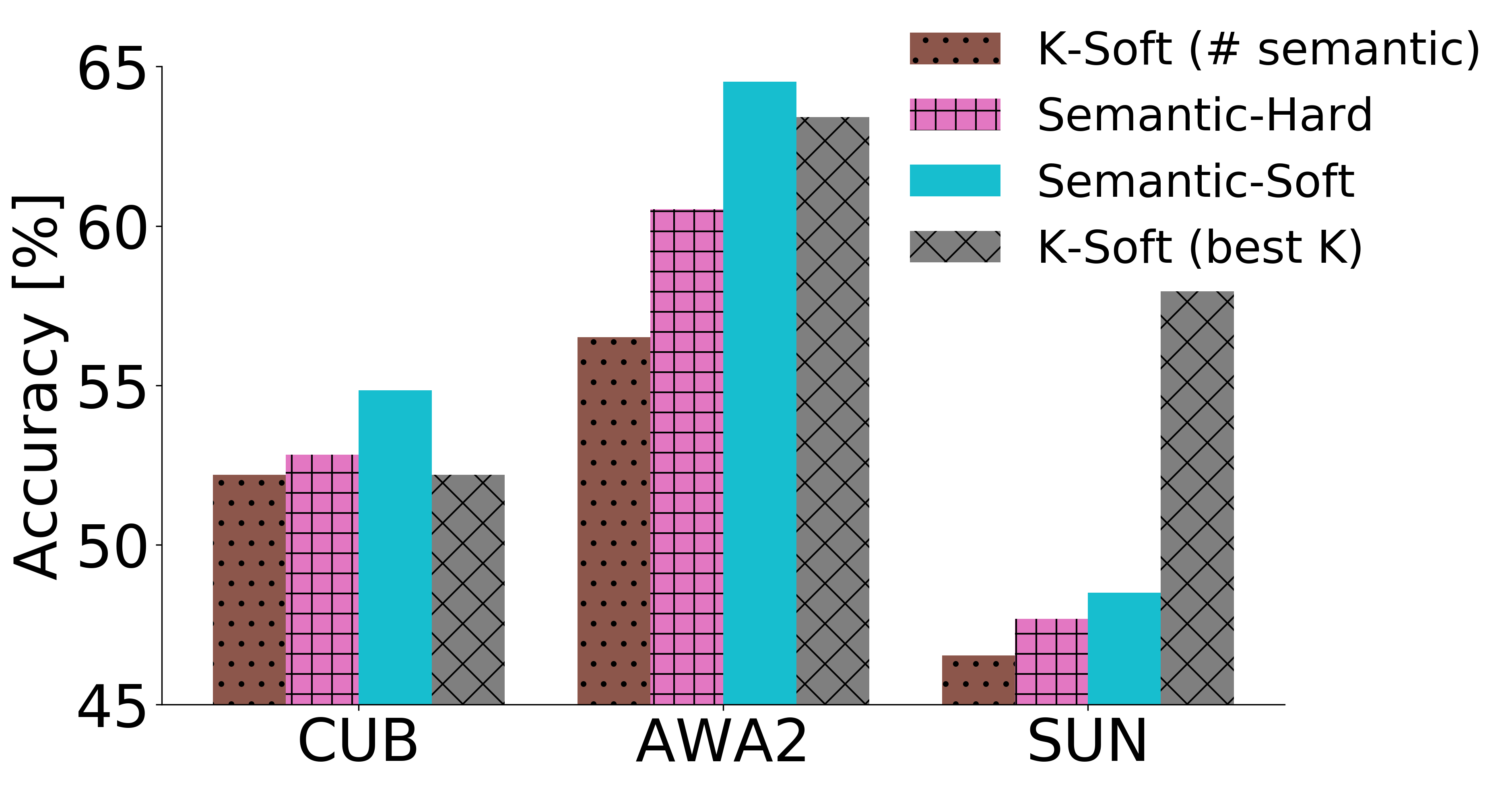} 
    \caption{Validation accuracy (in \%) of \lago{} variants for three ZSL benchmark datasets. (1)  On all the datasets, using semantic grouping information improves the performance relative to \lagoK{} with K = number of semantic hard-groups. (2) We used these results to select which variant to use for the test set. Explicitly, when training the model on train+validation data, we used the \textit{\SemanticSG{}} variant for CUB \& AWA2, and \textit{\lagoK{}} variant for SUN.}
    \label{fig_semantic_gain_val}
\vspace{-10pt}
\end{figure}

\begin{table}[t]
    \vskip 0.15in
    \begin{center}
      \begin{small}\begin{sc}
    \begin{tabular}{lccc}
    \toprule
    {} &   \textbf{CUB} &  \textbf{AWA2} &   \textbf{SUN} \\
    \midrule
    \textbf{DAP  }  &           40.0 &           46.2 &           39.9 \\
    \textbf{ALE  }  &           54.9 &           62.5 &  \textbf{58.1} \\
    \textbf{ESZSL}  &           53.9 &           58.6 &           54.5 \\
    \textbf{SYNC }  &           55.6 &           46.6 &           56.3 \\
    \textbf{SJE   } &           53.9 &           61.9 &           53.7 \\
    \textbf{DEVISE} &           52.0 &           59.7 &           56.5 \\
    \textbf{ZHANG2018*} &      48.7-57.1 &  58.3-\textbf{70.5} &      57.8-\textbf{61.7} \\
    
    \midrule
    \textbf{LAGO (ours)} &  \textbf{57.8 } &  \textbf{64.8 } &           57.5 \\
    \bottomrule
    \end{tabular}

      \end{sc}\end{small}
    \end{center}
    \caption{Test accuracy (in \%) of \lago{} and compared methods on three ZSL benchmark datasets. 
    We follow the protocol of a meta-analysis by \citep{xianCVPR} and compare to leading methods evaluated with it.
    Only one LAGO variant is shown, selected using a validation set.  See Table \ref{table_results_all_variants} in the supplementary for more results. \lago{} outperforms previous baselines on CUB and AWA2 by a significant margin. On SUN, \lago{} loses by a small margin. (*) Comparison with {\em Zhang2018} is inconclusive. \citet{zhang_kernel} report the results for 7 kernel types on the test set, but results on a validation set were not published, hence taking the best kernel over the test set may be  optimistic.
    }
    \label{table_results}
    \vspace{-15pt}
\end{table}

\paragraph{Design decisions:}
\textbf{(1)} We use a uniform prior for $\prob(z)$ as in \citep{xianCVPR,DAP,ESZSL}. $\prob(\am)$ can be estimated by marginalizing over $\prob(\am,z)$, but as in ESZSL and DAP, we found that uniform priors performed better empirically.
\textbf{(2)} To approximate the complementary attribute terms we used a De-Morgan based approximation for $\prob(\ack=T|z) \approx \prod_{m \in \Gk \backslash \{\ack\}} \prob(a_m^c|z)$. For $\prob(\ack=T|\x)$, we found that setting a constant value was empirically better than using a De-Morgan based approximation. 
\textbf{(3)} Our model does not use an explicit supervision signal for learning the weights of the attributes-prediction layer. Experiments showed that usage of explicit attributes supervision, by setting a non-zero value for $\alpha$ in \eqref{eq_loss}, results in deteriorated performance.

In Section \ref{sec_ablation}, we demonstrate the above design decision with ablation experiments on the validation sets of CUB and AWA2.

\paragraph{Implementation and training details:} The Supplementary Material (\ref{sec_model_training_supp}) describes the training protocol, including the  cross-validation procedure, optimization and tuning of hyper parameters. 

\vspace{-5pt}
\subsection{RESULTS}
\vspace{-5pt}
\label{sec_results}
Our experiments first compare variants of LAGO, and then compare the best variant to baseline methods. We then study in more depth the properties of the learned models.


Figure \ref{fig_acc_vs_K_on_val} shows validation-set accuracy of \lagoK{} variants as a function of the number of groups ($K$) and for the \Singletons{} baseline. We used these results to select the optimal number of groups K. In these experiments, even-though no prior information is provided about grouping, LAGO successfully learns group assignments from data, performing better than the group-naive \Singletons{} baseline. The performance degrades largely when the number of groups is small.

Figure \ref{fig_semantic_gain_val} shows validation-set accuracy for main variants of \lago{} for three benchmark datasets. We used these results to select the variant of \lago{} applied to the test split of each dataset. Specifcially, when training the model on train+validation data, we used \textit{\SemanticSG{}} for CUB \& AWA2, and \textit{\lagoK{}} ($K=40$) for SUN. This demonstrates that \lago{} is useful even when the semantic-grouping prior is of low quality as in SUN. Figure \ref{fig_semantic_gain_val} also shows that semantic grouping, significantly improves performance, relative to \lagoK{} with a similar number of groups.

We draw three conclusions from Figures \ref{fig_acc_vs_K_on_val}-\ref{fig_semantic_gain_val}. (1) The prior grouping based on attribute semantics contains very valuable information that LAGO can effectively use. (2) \lago{} succeeds even when no prior group information is given, effectively learning group assignments from data. (3) Using the semantic hard-groups as a prior, allows us to soften the semantic hard-groups and optimize the grouping structure from data.

Table \ref{table_results} details the main empirical results, comparing test accuracy of \lago{} with the competing methods. Importantly, to guarantee "apple-to-apple" comparison, evaluations are made based on the standard evaluation protocol from \citet{xianCVPR}, using the same underlying image features, data splits and metrics. Results are averaged over 5 random initializations (seeds) of the model weights ($W, \vvv$). Standard-error-of-the-mean (S.E.M) is $\tildeapprox$0.4\%.
On CUB and AWA2, \lago{} outperform all competing approaches by a significant margin. On CUB, reaching 57.8\% versus 55.6\% for SYNC \citep{SYNC}. On AWA2, reaching 64.8\% versus 62.5\% for ALE \citep{ALE}. On SUN, \lago{} loses by a small margin (57.5\% versus 58.1\%). Note that comparison with "Zhang2018" \citep{zhang_kernel} is inconclusive. "Zhang2018" reports the results for 7 kernel types on the test set, but results on a validation set were not published, hence taking the best kernel over the test set may be optimistic.

\paragraph{\Singletons{} versus DAP:}
\Singletons{} is a reminiscent of DAP, but unlike DAP, it applies a soft relaxation that balances between appearance of an attribute and its negation. Interestingly, this minor change allows \Singletons{} to outperform DAP by $\tildeapprox \text{40\%}$ on average over all three datasets, while keeping an appealing simplicity as of DAP (Supplementary Table \ref{table_results_all_variants}). 
\vspace{-10pt}

\ignore{ 
\vspace{-10pt}
\paragraph{\lago{} with few groups:}
When the number of groups is small, the accuracy of LAGO is poor (Fig \ref{fig_acc_vs_K_on_val}, \ref{fig_semantic_gain_val}), because LAGO becomes too permissive. E.g. in the 1 group case, an OR is applied over all attributes and no AND, leading to spurious matches. Details in Supplementary \ref{sec_lago_few_groups_supp}.
}

\paragraph{\lago{} with few groups:}
When the number of groups is small, the accuracy of LAGO is poor (Fig \ref{fig_acc_vs_K_on_val}, \ref{fig_semantic_gain_val})
This happens because when groups have too many attributes, the AND-OR structure of LAGO becomes too permissive. For example, when all attributes are grouped into a single group, an OR is applied over all attributes and no AND, leading to many spurious matches when partial attributes are observed. A similar effect is observed when applying LAGO to SUN data which has only 4 semantic hard groups for 102 attributes. Indeed applying \Semantic{} to SUN performs poorly since it is too permissive. Another interesting effect arises when comparing the the poor performance of the single-group case with ESZSL. ESZSL is convex and with a closed-form solution, hence reaching the optimum is guaranteed. Single-group LAGO is non-convex (due to sigmoidal activation) making  it harder to find the optimum. Indeed, we observed a worse training accuracy for single-group LAGO compared with ESZSL (61\% vs 84\% on CUB), suggesting that single-group LAGO tends to underfit the data.

\vspace{-10pt}
\paragraph{Learned Soft Group Assignments $\Gamma$:}
We analyzed the structure of learned soft group assignments ($\Gamma$) for \lagoK{} (details in Supplementary \ref{sec_gamma_analysis_supp}). We found two interesting observations: First, we find that the learned $\Gamma$ tends to be sparse: with 2.5\% non-zero values on SUN, 8.7\% on AWA2 and 3.3\% on CUB. Second, we observed that the model tends to group anti-correlated attributes. This is consistent with human-based grouping, whose attribute are also often anti correlated (red foot, blue foot). In SUN, 45\% of attribute-pairs that are grouped together were anti-correlated, versus 23\% of all attribute-pairs. In AWA2, 38\% vs 5\% baseline, CUB 16\% vs 10\% baseline (p-value $\!\leq\!$ 0.003, KS-test).

\begin{figure*}[t]
    \centering
  \includegraphics[width=16cm, 
  trim={0.cm 0.cm 0.cm 0.cm},
  clip]{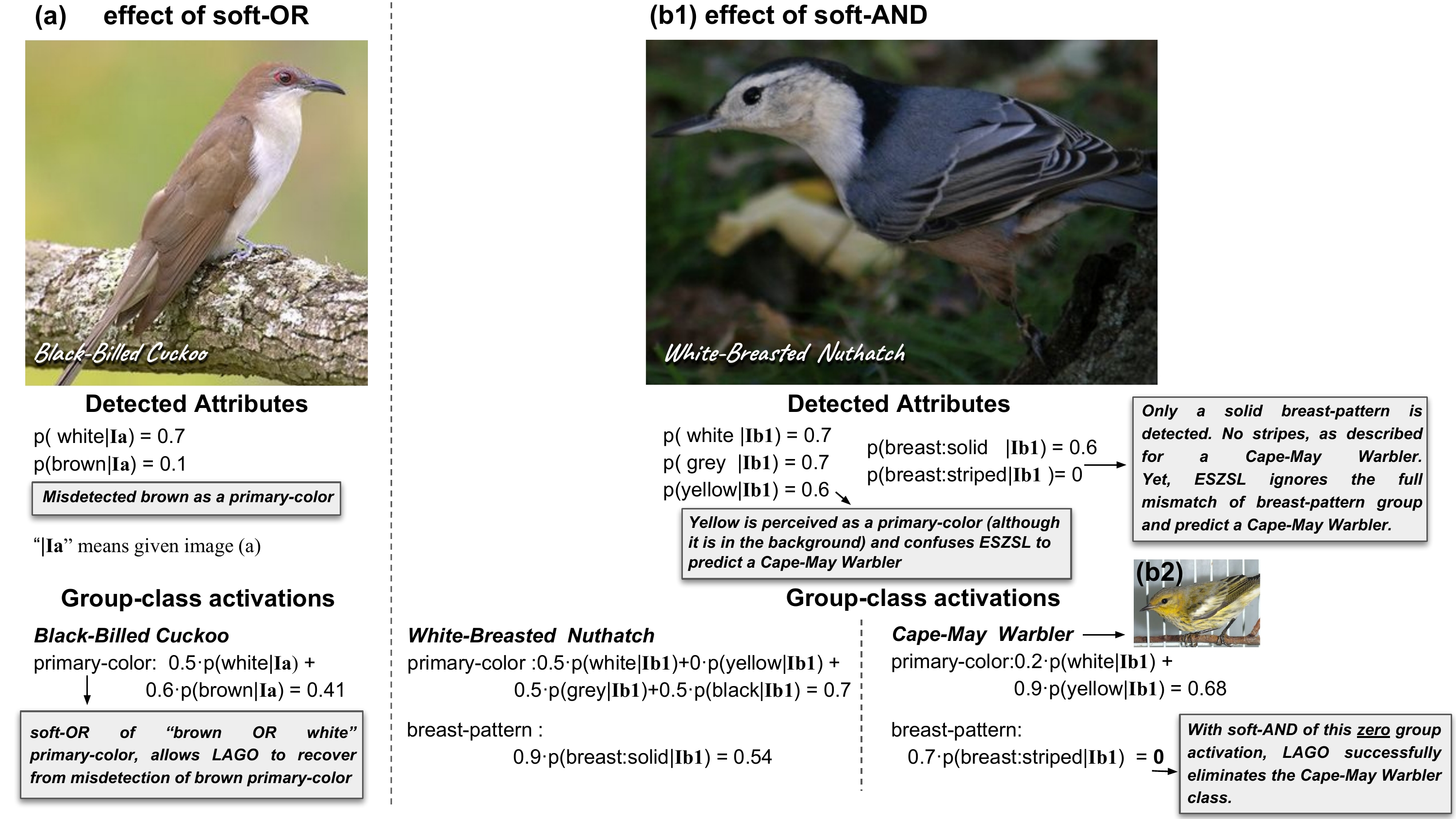} 
    \caption{Qualitative examples. 
    \textbf{(a)} LAGO correctly classifies a Black-billed Cuckoo, due to soft-OR of ``brown or white'' primary-color, although a detector misses its brown primary-color.
\textbf{(b1)} LAGO correctly classifies a White-Breasted Nuthatch: The soft-intersection across groups prevents incorrect classification. ESZSL incorrectly classified Image (b1) as
class of (b2), despite irrelevant attribute groups like its breast pattern
\textbf{(b2)} A typical Cape-May  Warbler. ESZSL mistakes (b1) image for this class 
 }
    \label{fig_qualitative}
\vspace{-10pt}
\end{figure*}

\vspace{-10pt}
\paragraph{Qualitative Results:}
To gain insight into why and when attribute grouping can reduce false positives and false negatives, we discuss in more depth two examples shown on Figure \ref{fig_qualitative}, predicted by \Semantic{} on CUB. The analysis demonstrates an interpretable quality of LAGO, allowing to "look under the hood" and explain class-predictions based on seen attributes.

\textbf{The effect of within-group disjunction (OR):} Image \ref{fig_qualitative}a is correctly classified by \lago{} as a \textit{Black-billed Cuckoo}, even-though a detector misses its brown primary-color. In more detail, for this class, raters disagreed whether the primary-color is mostly brown ($p(brown|z) = 0.6$) or white ($0.5$), because this property largely depends on the point-of view. Somewhat surprisingly, the primary color in this photo was detected to be mostly white ($\prob(white|\x)=0.7$), and hardly brown ($0.1$), perhaps because of a brown branch that interferes with segmenting out the bird. Missing the brown color hurts any classifier that requires both brown and white, like DAP. LAGO treats the detected primary color as a good match because it takes a soft OR relation over the two primary colors, hence avoids missing the right class. 

\textbf{The effect of group conjunction (AND):} Image \ref{fig_qualitative}b.1 was correctly classified by LAGO as a \textit{White-Breasted Nuthatch}, even-though a detector incorrectly detects a yellow primary color $(\prob(yellow|\x) = 0.6)$ together with white and grey primary colors ($0.7$). As a comparison, the perceived yellow primary color confused ESZSL to mistake this image for a \textit{Cape-May Warbler}, shown in image (b.2). Since ESZSL treats attributes as "flat", it does not use the fact that the breast pattern does not match a Warbler, and adheres to other attributes that produce a false positive detection of the Warbler. Yet, LAGO successfully avoids being confused by the yellow primary color, since the Nuthatch is expected to have a solid breast pattern, which is correctly detected  $\prob(breast:solid|\x)=0.6$. The Warbler is ranked lower because it is expected to have a striped breast pattern, which does not satisfy the AND condition because stripes are not detected $p(breast:striped|x)=0$.

\vspace{-5pt}
\subsection{ABLATION EXPERIMENTS}
\vspace{-5pt}
\label{sec_ablation}

We carried empirical ablation experiments with the semantic hard-grouping of \lago{}. Specifically, we tested three design decisions we made, as described above.  {\bf (1) Uniform} relates to taking a uniform prior for $\prob(\am)$, which is the average of the estimated $\prob(\am)$. ``Per-attribute'' relates to using the estimated $\prob(\am)$  directly. {\bf (2) Const} relates to setting a constant value for the approximation of the complementary attribute $\prob(\ack|\x)$. ``DeMorgan'' relates to approximating it from predictions of other attributes with De-Morgan's rule. {\bf{ (3) Implicit}} relates to setting a zero weight ($\alpha=0$) for the loss term of the attribute supervision. I.e. attributes are learned implicitly, because only class-level super vision is given. ``Explicit'' related to setting a non-zero $\alpha$ respectively. 

Table \ref{table_ablation} (in Supplementary) shows contributions of each combination of the design decisions to prediction accuracy, on the validation set of CUB and AWA2. The results are consistent for both CUB and AWA2. The most major effect is contributed for the uniform prior of $\prob(\am)$. All experiments that use the uniform prior yield better accuracy. We observe that taking a uniform prior also reduces  variability due to the other approximations we take. Specifically, on CUB there is $\approx$ 4.5\% best-to-worst gap with a uniform prior, vs $\approx$ 12.5\% without ($\approx$ 11\% vs $\approx$ 16\% for AWA2 respectively). Next we observe that in the uniform case, approximating $\prob(\ack|\x)$ by a constant, is superior to approximating it with De-Morgan's rule, and similarly, reduces the impact  of the variability of the implicit/explicit condition. Last, the contribution of attributes supervision condition mostly depends on selection of the previous two conditions. 

\vspace{-5pt}
\section{DISCUSSION}
\vspace{-5pt}

Three interesting future research directions can be followed. First, since  \lago{} is probabilistic, one can plug measures for model uncertainty \citep{YGalUncertainty}, to improve model prediction and increase robustness to adversarial attacks.  Second, descriptions of fine-grained categories often provide richer logical expressions to describe and differentiate classes.  It will be interesting to study how \lago{} may be extended to incorporate richer relations that could be explicitly discriminative \citep{RamaCVPR}. For example, Wikipedia describes \textit{ White-Breasted-Nuthatch} to make it distinct from other, commonly confused, Nuthatches by:  ``\textit{Three other, significantly smaller, nuthatches have ranges which overlap that of white-breasted, but none has white plumage completely surrounding the eye. }''. 

Third, when people describe classes, they often use a handful of attributes instead of listing all values for the full set of attributes. The complementary attribute used in \lago{} allows to model a ``don't-care'' about a group, when no description is provided for a group. Such an approach could enable to recognize visual entities based on a handful and partial indication of semantic properties.
 
\vspace{-5pt}
\section{SUMMARY}
\vspace{-5pt}
We presented \lago{}, a new probabilistic zero-shot-learning approach that can be trained end-to-end. \lago{} approximates $\prob(class\!=\!z|image\!=\!\x)$ by capturing natural soft {\em and-or} logical relations among \textit{groups of attributes}, 
unlike most ZSL approaches that represent attributes as embedded in a ``flat'' space. LAGO learns the grouping structure from data, and can effectively incorporate prior domain knowledge about the grouping of attributes when available. We find that \lago{} achieves new state-of-the-art result on CUB \citep{CUB}, AWA2 \citep{DAP}, and is competitive on SUN \citep{SUN}. 
Finally, \lago{} reveals an interesting unified formulation over seemingly-unrelated ZSL approaches, DAP \citep{DAP} and ESZSL \citep{ESZSL}.

\clearpage
\newpage
{\small
\bibliographystyle{iclr2018_conference}
\bibliography{egbib}
}

\section*{SUPPLEMENTARY MATERIAL}

\renewcommand\thefigure{A.\arabic{figure}}    
\renewcommand\thetable{A.\arabic{table}}   
\renewcommand\thesection{\Alph{section}}   
\renewcommand{\theequation}{A.\arabic{equation}}
\setcounter{figure}{0}  
\setcounter{table}{0}  
\setcounter{section}{0}  

\section{IMPLEMENTATION AND TRAINING DETAILS}
\label{sec_model_training_supp}
The weights $W$ were initialized with orthogonal initialization \citep{saxe2013exact}. The loss in \eqref{eq_loss} was optimized with Adam optimizer \citep{ADAM}. We used cross-validation to tune early stopping and hyper-parameters. When the learning rate is too high, the number of epochs for early-stopping varies largely with the weight seed. Therefore, we chose a learning rate that shows convergence within at least 40 epochs. Learning rate was searched in [3e-6, 1e-5, 3e-5, 1e-4, 3e-4]. From the top performing hyper-parameters, we chose the best one based on an average of additional 3 different seeds. Number-of-epochs for early stopping, was based on their average learning curve. For $\beta, \lambda$,  L2 regularization params, we searched in [0, 1e-8, .., 1e-3]. 

For learning soft groups, we also tuned the learning rate of $\vvv$ in [0.01, 0.1, 1], of $\zeta$ in [1, 3, 10], and when applicable, the number of groups $K$ in [1, 10, 20, 30, 40, 60], or semantic prior $\psi$ in [1e-5, .., 1e-2]. We tuned these hyper-params by first taking a coarse random search, and then further searching around the best performing values.

To comply with the mutual-exclusion approximation (\ref{eq_disj_mux}), if the group sum  $\sum_{m \in \Gk}\prob(\am=T|z)$ is larger than 1, we normalize it to 1. We do not normalize if the sum is smaller than 1 in order to allow \lago{} to account for the complementary case. We apply this normalization only for the \lago{}-Semantic variants, where a prior knowledge about grouping is given.

After selecting hyper-parameters with cross-validation, models were retrained on both the training and the validation classes.

\subsection{EVALUTATION METRIC}
\label{sec_acc_metric_supp}

We follow \citet{xianCVPR} and use a class-balanced accuracy metric which averages correct predictions independently per-class before calculating the mean value:
\vspace{-5pt}
\begin{equation}
    acc_{\Z} = \frac{1}{|\Z|}\sum_{z=1}^{|\Z|}
    \frac{\text{\# of correct predictions in \,} z}{\text{\# of samples in \,} z} .
\end{equation}

\section{LEARNED SOFT-GROUP ASSIGNMENTS \texorpdfstring{($\Gamma$)}:}
\label{sec_gamma_analysis_supp}
We analyzed the structure of learned soft group assignments ($\Gmk = \prob({m \in \Gk})$) for \lagoK{}, initialized by a uniform prior. We found two types of interesting structures:

First, we find that the learned $\Gamma$ tends to be sparse: with 2.5\% non-zero values on SUN, 8.7\% on AWA2 and 3.3\% on CUB. As a result, the learned model has small groups, each with only a few attributes. Specifically, $\Gamma$ maps each attribute to only a single group on SUN  (K=40 groups) and CUB (K=30), and to 2-3 groups on AWA2  (K=30 groups).

Second, we tested which attributes tend to be grouped together, and found that the model tends to group anti-correlated attributes. To do this, we first quantified for each pair of attributes, how often they tend to appear together in the data. Specifically, we estimated the occurrence pearson-correlation for each pair of attributes across samples (CUB, SUN) or classes (AWA2). Second, we computed the grouping similarity of two attributes as the inner product of their corresponding rows in Gamma, and considered an attribute pair to be grouped together if this product was positive (note that rows are very sparse).
Using these two measures, we observed that the model tends to group anti-correlated attributes. This is consistent with human-based grouping, whose attribute are also often anti correlated (red foot, blue foot). In SUN, 45\% of attribute-pairs that are grouped together were anti-correlated, compared to  23\% of the full set of pairs. (AWA2 38\% vs 5\% baseline, CUB 16\% vs 10\% baseline). These differences were also highly significant statistically (Kolmogorov-Smirnov test p-value $<$ 3e-3)

\section{ROBUSTNESS TO NOISE}
\label{sec_noise_robust_supp}
\begin{figure}[t]
    \centering
    \includegraphics[width=7cm, trim={0cm 0cm 0cm 0cm},clip]{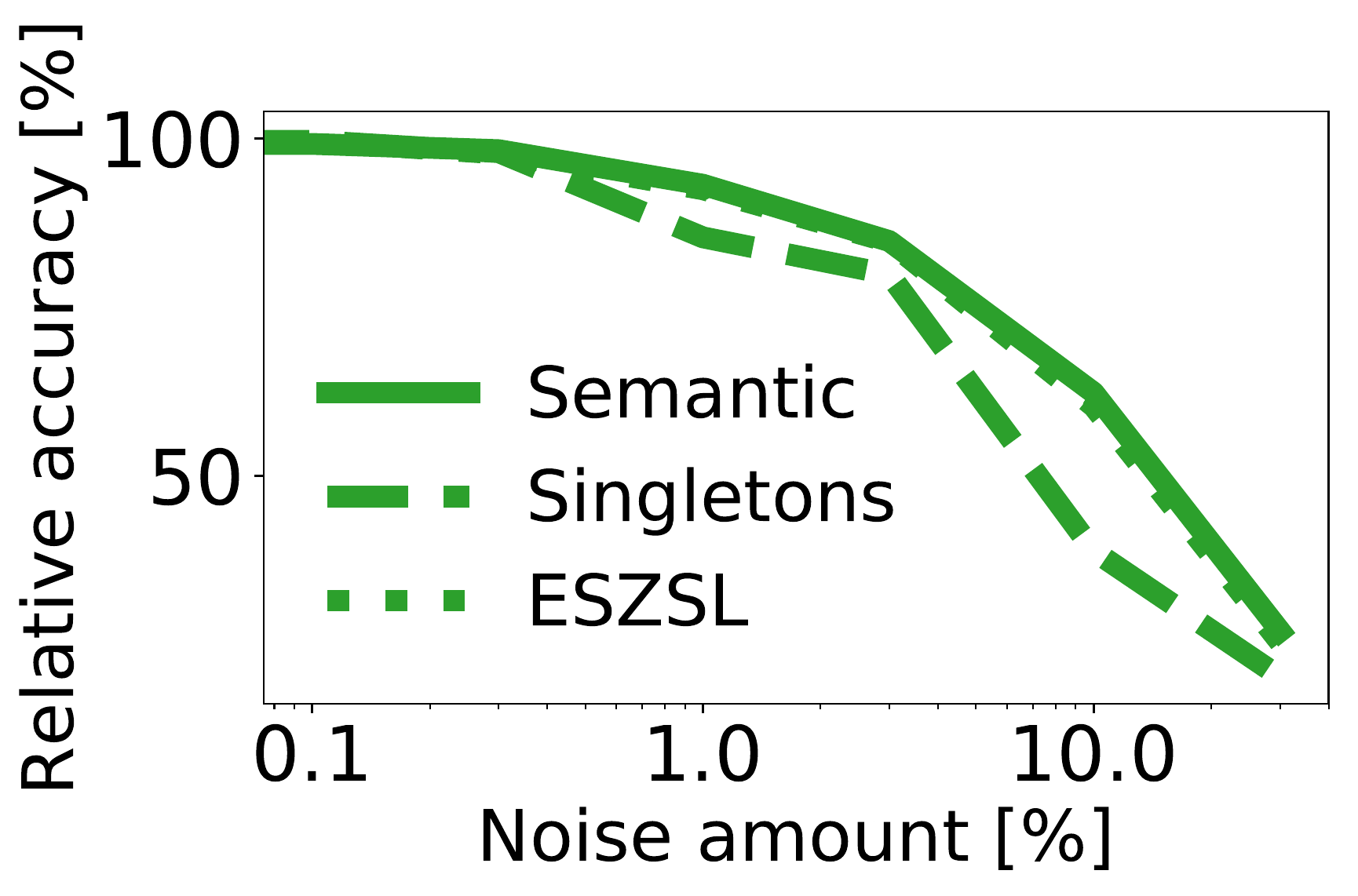} 
    \caption{Robustness to salt \& pepper noise.  The relative accuracy on CUB of three models as a function of ration of injected noise to class-level description $\prob(\am|z)$. Values are averages over 5 noise-seeds. \Semantic{} and ESZSL show a similar sensitivity to noise, while \Singletons{} is more sensitive due to its all-AND structure. The relative accuracy is calculated against each model own zero-noise baseline.}
    \label{fig_noise_sensitivity}
\vspace{-10pt}
\end{figure}

We tested \Semantic{}, \Singletons{} and ESZSL with various amount of salt \& pepper noise (Figure \ref{fig_noise_sensitivity}) injected to class-level description $\prob(\am|z)$ of CUB. While \Semantic{} and ESZSL show a similar sensitivity to noise, \Singletons{} is more sensitive due to its all-AND structure.

\begin{table*}[t]
    \vskip 0.15in
    \begin{center}
      \begin{small}\begin{sc}
\begin{tabular}{lccc}
\toprule
{} &   \textbf{CUB} &  \textbf{AWA2} &   \textbf{SUN} \\
\midrule
\textbf{DAP  } &           40.0 &           46.2 &           39.9 \\
\textbf{ALE  } &           54.9 &           62.5 &  \textbf{58.1} \\
\textbf{ESZSL} &           53.9 &           58.6 &           54.5 \\
\textbf{SYNC } &           55.6 &           46.6 &           56.3 \\
\textbf{SJE   } &           53.9 &           61.9 &           53.7 \\
\textbf{DEVISE} &           52.0 &           59.7 &           56.5 \\
\textbf{ZHANG2018 } &      48.7 - 57.1     &           58.3-70.5 &           57.8-61.7 \\
\midrule
\textbf{\Singletons{}} &           54.5 &           63.7 &           57.3 \\
\textbf{\lagoK{}}     &           55.3 &           59.7 &  57.5 \\
\textbf{\Semantic{}}   &  58.3 &           60.4 &           47.1 \\
\textbf{\SemanticSG{}} &           57.8 &  64.8 &           48.0 \\
\midrule
\textbf{\lago{} (cross-validation)} &  \textbf{57.8} &  \textbf{64.8} &           57.5 \\
\bottomrule
\end{tabular}
      \end{sc}\end{small}
    \end{center}
    \caption{Test accuracy for all the variants of \lago{} on three benchmark datasets,  averaged over 5 random initializations of model weights. Standard-error-of-the-mean (S.E.M) is $\tildeapprox$ 0.1\% for the hard groups variants and $\tildeapprox$ 0.4\% for the soft-groups variants.}
    \label{table_results_all_variants}
\end{table*}

\begin{table*}[ht]
        \vskip 0.15in
    \begin{center}
      \begin{small}\begin{sc}
    \begin{tabular}{lllll}
    \toprule
           $\prob(\am)$ & $P(\ack|\x)$ & Attributes supervision &             CUB &            AWA2 \\
    \midrule
           Uniform &   Const & Implicit & \textbf{52.85} & \textbf{60.53} \\
           Uniform &   Const & Explicit &          52.17 &          60.17 \\
           Uniform &DeMorgan & Explicit &          51.75 &          57.93 \\
           Uniform &DeMorgan & Implicit &          48.41 &          49.54 \\
     Per-attribute &DeMorgan & Explicit &          47.71 &          53.32 \\
     Per-attribute &   Const & Explicit &          42.68 &          52.05 \\
     Per-attribute &   Const & Implicit &          39.31 &          51.88 \\
     Per-attribute &DeMorgan & Implicit &           35.3 &          37.21 \\
    \bottomrule
    \end{tabular}
      \end{sc}\end{small}
    \end{center}
    \caption{Ablation experiments: Validation accuracy (in \%) for CUB and AWA2, for combinations of model-design variants, with the semantic hard-grouping of \lago{}. Results are given in descending order based on CUB. {\em Uniform} vs {\em Per-attribute} relates to taking a uniform prior for $\prob(\am)$. {\em Const}  {\em DeMorgan} relates to setting a constant value for approximating the complementary attribute $\prob(\ack|\x)$ vs an approximation derived by De-Morgan's rule. {\em Implicit} vs {\em Explicit} relates to setting a zero weight ($\alpha=0$) for the loss term of the attribute supervision. Namely, attributes are learned implicitly, since only class-level super vision is given. The uniform prior on $\prob(\am)$ has the largest impact, second is the usage of a constant value for $\prob(\ack|\x)$, and the last relates to nulling the attribute supervision loss. See details on Section \ref{sec_ablation}}
    \label{table_ablation}
\end{table*}

\section{DETAILED DERIVATION}
\subsection{$\prob(\am|\gkz\!\!=\!\!T)$ EQUALS $\prob(\am|Z\!\!=\!\!z)$}
\label{sec_am_cond_gkz_supp}
Here we explain why \eqref{eq_am_cond_gkz} below is true.
\begin{equation}
    \label{eq_am_cond_gkz}
    \prob(\am|\gkz\!\!=\!\!T)=\prob(\am|Z\!\!=\!\!z), 
\end{equation}

It is based on the definition of $\gkz$: $\gkz$ is the classifier of $z$ based on $\a_k$. Therefore $p(\gkz|\a_k)\!\!=\!\!p(z|\a_k)$, and by marginalization we get: (*) $\prob(\gkz\!=\!T, \am)\!\!=\!\!\prob(z,\am)$,  (**) $\prob(\gkz\!=\!T)\!\!=\!\!\prob(Z\!=\!z)$. Next, using conditional probability chain rule on (*), yields
\begin{equation}
    \label{gkz_am_chained}
    \prob(\am|\gkz\!=\!T)\prob(\gkz\!=\!T) = \prob(\am|Z=z)\prob(Z=z) \quad.
\end{equation}
Then, (**) transforms \brackref{gkz_am_chained} to the required equality:
\begin{equation}
    \label{am_cond_gkz}
    \prob(\am|\gkz\!=\!T) = \prob(\am|Z=z) \quad.
\end{equation}

Intuitively, the right side of \brackref{am_cond_gkz}, is the probability of observing $\am$ for a class $z$, like $\prob(stripes|zebra)$. This is the same probability of observing the attribute given the class while focusing on its respective group, namely $\prob(\am\!=\!T|\gkz\!=\!T)\!=\!\prob(stripes|\textit{focus on zebra pattern})$.

\subsection{DERIVATION OF GROUP CONJUNCTION:} 
\label{sec_derivation_eq_conj_supp}
This derivation is same as in DAP (Lampert 2009), except we apply it at the group level rather than the attribute level. We denote ${g_{1,z} \dots g_{\K,z}}$ by $\gz$ and approximate the following combinatorially large sum:
\begin{equation}
    \label{eq_groups_marginals}
    \prob(Z\!=\!z|\x)\!=\!
    \sum_{\gz\in \{T,F\}^K} 
    \prob(Z\!=\!z|\gz)\prob(\gz|\x) \quad.
\end{equation}
First, using Bayes \brackref{eq_groups_marginals} becomes
\begin{equation}
    \label{eq_groups_marginals_bayes}
    \sum_{\gz\in \{T,F\}^K} 
    \frac{\prob(\gz|Z\!=\!z)\prob(Z\!=\!z)}{\prob(\gz)}\prob(\gz|\x)
\end{equation}
Second, we approximate $\prob(\gz|Z\!=\!z)$ to be
\begin{equation}
   \prob(\gz|Z\!=\!z) = 
    \begin{cases}
      1, & \text{if}\ g_{1,z}\!=\!T \dots g_{\K,z}\!=\!T \\
      0, & \text{otherwise}
    \end{cases}
\end{equation}
which transforms \brackref{eq_groups_marginals_bayes} to
\begin{equation}
    \label{eq_groups_approx}
    \prob(Z\!=\!z|\x)\!\approx \prob(Z\!=\!z)\frac{\prob(g_{1,z}\!=\!T \dots g_{\K,z}\!=\!T|\x)}{\prob(g_{1,z}\!=\!T \dots g_{\K,z}\!=\!T)}
\end{equation}

Third, we approximate the numerator of \brackref{eq_groups_approx} with the assumption of conditional independence of groups given an image (by observing an image we can judge each group independently), 
\begin{equation}
\label{eq_numerator_groups_approx}
    \prob(g_{1,z}\!=\!T \dots g_{\K,z}\!=\!T|\x) \approx 
    \prod\limits_{k=1}^K \prob(\gkz\!=\!T|\x) 
\end{equation}

Fourth, we approximate the denominator of \brackref{eq_groups_approx} to its factored form $\prob(g_{1,z}\!=\!T \dots g_{\K,z}\!=\!T) \approx \prod\limits_{k=1}^K {\prob(\gkz\!=\!T)}$, and with \brackref{eq_numerator_groups_approx} we arrive at: 
\begin{equation}
\label{eq_conj}
    \prob(Z\!=\!z|\x) \approx \prob(Z\!=\!z)\prod\limits_{k=1}^K \frac{\prob(\gkz\!=\!T|\x)}{\prob(\gkz\!=\!T)} \quad.
\end{equation}

\subsection{A DERIVATION OF SOFT GROUP MODEL}
\label{sec_derviation_soft_groups_supp}
Here we adapt \lago{} to account for soft group-assignments for attributes, by extending the within-group part of the model. We start with partitioning $\prob(\gkz\!=\!T|\x)$ to a union (OR) of its contributions, repeated below for convenience, 
\begin{multline}
\prob(\gkz|\x) = \\ \prob(\gkz, \medcup\limits_{m \in \Gk} a_m=T|\x) +  \prob(\gkz,\ack=T |\x),
\end{multline}
and instead, treat the attribute-to-group assignment ${(m \in \Gk)}$, as a probabilistic assignment, yielding:
\begin{multline}
\label{eq_disj_softer}
\prob(\gkz|\x) = \\ \prob(\gkz, \medcup\limits_{m = 1}^{|\A|} ({m \in \Gk}, a_m=T|\x)) +  \prob(\gkz,\ack=T |\x),
\end{multline}
Note that the attribute-to-group assignment $(m \in \Gk)$ is independent of the current given image $\x$, class $z$ or the True / False occurrence of an attribute $\am$. Repeating the mutual exclusion approximation (\ref{eq_disj_mux}) yields,
\begin{equation}
\prob(\gkz|\x) \approx \sum_{m = 1}^{|\A|} \prob(\gkz, {m \in \Gk}, a_m=T|\x).
\end{equation}
Using the independence of $({m \in \Gk})$, yields
\begin{equation}
\prob(\gkz|\x) \approx \sum_{m = 1}^{|\A|} \prob({m \in \Gk})\prob(\gkz, a_m=T|\x).
\end{equation}
Defining $\Gmk = \prob({m \in \Gk})$, yields:
\begin{equation}
\prob(\gkz|\x) \approx \sum_{m = 1}^{|\A|} \Gmk \prob(\gkz, a_m=T|\x).
\end{equation}
As in section \ref{sec_factor}, using the Markov chain property $\X \rightarrow \A \rightarrow \G$ and $\prob(\gkz = T |\am) = \frac{\prob(\am|z)\prob(\gkz = T)}{\prob(\am)}$ results with \eqref{eq_soft_gkz_cond_x}, repeated below:
\begin{multline}
    \prob(\gkz=T|\x) {} \approx \\ \prob(\gkz=T)\sum_{m = 1}^{|\A|} \Gmk\frac{\prob(\am=T|z) }{\prob(\am=T)} \prob(  \am=T | \x)
\end{multline}

\subsubsection{APPROXIMATING THE COMPLEMENTARY TERM}
With soft groups, the complementary term is defined as
\begin{equation}
 \ack = \Big(\medcup\limits_{m = 1}^{|\A|} ({m \in \Gk}, a_m=T|\x))\Big)^c
\end{equation}
To approximate $\prob(\ack=T|z)$ we can use De-Morgan's rule over a factored joint conditional probability of group-attributes. I.e. 
\begin{multline}
\prob(\ack=T|z) \approx \prod_{m = 1}^{|\A|}( 1 - \prob({m \in \Gk}, \am=T|z)) = \\  \prod_{m = 1}^{|\A|} (1-\Gmk \prob(\am=T|z)),
\end{multline}
where the latter term is derived by the independence of $({m \in \Gk})$

\subsection{DAP, ESZSL AS SPECIAL CASES OF \lago{}}
\label{sec_dap_eszsl_supp}
Two extreme cases of \lago{} are of special interest: having each attribute in its own singleton group ($K=|A|$), and having one big group over all attributes ($K=1$).

Consider first assigning each single attribute $\am$ to its own singleton group ($K=|A|$ and $m=k$). We remind that we defined $\Gk' = \Gk \medcup \ack$. Therefore, $\Gk'$ has only two attributes $\{\am, \ack\}$, which turns the sum in \eqref{eq_prodsum}, to a sum over those elements:
\begin{multline}
\label{eq_singleton_ack}
    \prob(z|\x) {} = \,\,\,\,\prob(z) \prod\limits_{k=1}^{\K}\Big[\frac{\prob(\am\!=\!T|z) }{\prob(\am\!=\!T)} \prob(\am\!\!=\!\!T|\x) + \\ \!\!\frac{\prob(\ack\!=\!T|z) }{\prob(\ack\!=\!T)} \prob(\ack\!\!=\!\!T|\x)  \Big] .
\end{multline}
In a singleton group, the complementary attribute $\ack$ becomes $\ack=\am^{c}$, and therefore \\ $\ack\!=\!T \Leftrightarrow \am\!=\!F$. This transforms \brackref{eq_singleton_ack} to:

\begin{multline}
    \label{soft_dap}
    \prob(z|\x) {} = \,\,\,\,\prob(z) \prod\limits_{m=1}^{|A|}\Big[ \frac{\prob(\am\!\!=\!\!T|z) }{\prob(\am\!\!=\!\!T)} \prob(\am\!\!=\!\!T|\x) + \\ \frac{\prob(\am\!\!=\!\!F|z) }{\prob(\am\!\!=\!\!F)} \prob(\am\!\!=\!\!F|\x) \Big] \quad.
\end{multline}
This formulation is closely related to DAP \citep{DAP}, where the expert annotation $\prob(\am\!\!=\!\!T|z)$ is thresholded to $\{0,1\}$ using the mean of the matrix $\uuu$ as a threshold, and denoted by $a^z_m$. Applying a similar threshold to \eqref{soft_dap} yields 
\begin{multline}
    \label{threshold_dap}    
    \prob(z|\x) {} = \prob(z) \prod\limits_{m=1}^{|A|}\Big[ \frac{a^z_m }{\prob(\am\!\!=\!\!T)} \prob(\am\!\!=\!\!T|\x) + \\ \frac{(1-a^z_m) }{\prob(\am\!\!=\!\!F)} \prob(\am\!\!=\!\!F|\x) \Big]
\end{multline}
Reducing \eqref{threshold_dap}, by taking only the cases where it is non-zero for its two parts, gives the posterior of DAP
\begin{equation}
    \label{dap}
    \prob(z|\x) {} = \\\prob(z) \prod\limits_{m=1}^{|A|} \frac{ \prob(\am\!\!=\!\!a^z_m|\x)}{\prob(\am\!\!=\!\!a^z_m)} \quad.
\end{equation}
This derivation reveals that in the extreme case of $K=|A|$ singleton groups, \lago{} becomes equivalent to a soft relaxation of DAP. 

At the second extreme, consider the case where all attributes are assigned to a single group, $K\!=\!1$. Taking a uniform prior for $\prob(z)$ and $\prob(\am)$, and writing $\prob(\am=T|\x)$ using the network model $\sigma(\x^\T W)$, transforms \eqref{eq_prodsum} to:
\begin{equation} 
    \prob(z|\x) \propto  \sum\limits_{m=1}^{|A|} \sigma(\x^\T W) \prob(\am=T|z).
\end{equation}
This can be viewed as a 2-layer architecture: First map image features to a representation in the attribute dimension, then map it to class scores by an inner product with the supervised entries of attributes-to-classes $\Smz=\prob(\am=T|z)$.
This formulation resembles ESZSL, which uses a closely related 2-layer architecture: $Score(z|\x) =\x^\T W U$, where $W$ first maps image features to a representation in the same attribute dimension, and then map it to class scores, with an inner product by the same attributes-to-classes entries $\Smz=\prob(\am=T|z)$.  \lago{} differs from ESZSL in two main ways: (1) The  attribute-layer in \lago{} uses a sigmoid-activation, while ESZSL uses a linear activation. (2)  \lago{} uses a cross-entropy loss, while ESZSL uses mean-squared-error. This allows ESZSL to have a closed-form solution where reaching the optimum is guaranteed.

This derivation reveals that at the extreme case of $K=1$, \lago{} can be viewed as a non-linear variant  that is closely related to ESZSL.

\end{document}